\documentclass[sigconf,nonacm]{acmart}
\AtBeginDocument{%
  }

\usepackage{multirow}

\begin{document}

%%
%% The "title" command has an optional parameter,
%% allowing the author to define a "short title" to be used in page headers.
\title{WebMall - A Multi-Shop Benchmark for Evaluating Web Agents}

%%
%% The "author" command and its associated commands are used to define
%% the authors and their affiliations.
%% Of note is the shared affiliation of the first two authors, and the
%% "authornote" and "authornotemark" commands
%% used to denote shared contribution to the research.
\author{Ralph Peeters}
\orcid{0000-0003-3174-2616}
\affiliation{%
  \institution{Data and Web Science Group}
  \institution{University of Mannheim}
  \city{Mannheim}
  \country{Germany}
}
\email{ralph.peeters@uni-mannheim.de}

\author{Aaron Steiner}
\orcid{0009-0006-6946-7057}
\affiliation{%
  \institution{Data and Web Science Group}
  \institution{University of Mannheim}
  \city{Mannheim}
  \country{Germany}
}
\email{aaron.steiner@uni-mannheim.de}

\author{Luca Schwarz}
\orcid{0009-0003-4751-4607}
\affiliation{%
  \institution{Data and Web Science Group}
  \institution{University of Mannheim}
  \city{Mannheim}
  \country{Germany}
}
\email{lucschwa@mail.uni-mannheim.de}

\author{Julian Yuya Caspary}
\orcid{0009-0001-0606-4115}
\affiliation{%
  \institution{Data and Web Science Group}
  \institution{University of Mannheim}
  \city{Mannheim}
  \country{Germany}
}
\email{jcaspary@mail.uni-mannheim.de}

\author{Christian Bizer}
\orcid{0000-0003-2367-0237}
\affiliation{%
  \institution{Data and Web Science Group}
  \institution{University of Mannheim}
  \city{Mannheim}
  \country{Germany}
}
\email{christian.bizer@uni-mannheim.de}

%%
%% By default, the full list of authors will be used in the page
%% headers. Often, this list is too long, and will overlap
%% other information printed in the page headers. This command allows
%% the author to define a more concise list
%% of authors' names for this purpose.
\renewcommand{\shortauthors}{Ralph Peeters, Aaron Steiner, Luca Schwarz, Julian Yuya Caspary, and Christian Bizer}

%%
%% The abstract is a short summary of the work to be presented in the
%% article.
\begin{abstract}
  LLM-based web agents have the potential to automate long-running web tasks, such as searching for products in multiple e-shops and subsequently ordering the cheapest products that meet the user’s needs. Benchmarks for evaluating web agents either require agents to perform tasks online using the live Web or offline using simulated environments, the latter allowing for the exact reproduction of the experimental setup. While DeepShop and ShoppingComp provide online benchmarks that require agents to perform challenging shopping tasks, existing offline benchmarks such as WebShop, WebArena, and Mind2Web cover only comparatively simple e-commerce tasks performed against a single shop containing product data from a single source. What is missing is an e-commerce benchmark that simulates multiple shops containing heterogeneous product data and requires agents to perform complex retrieval tasks. We fill this gap by introducing WebMall, the first offline multi-shop benchmark for evaluating web agents on challenging comparison shopping tasks. WebMall consists of four simulated shops populated with product data extracted from the Common Crawl. The WebMall tasks range from specific product searches and price comparisons to advanced searches for complementary or substitute products, as well as checkout processes. We validate WebMall using eight agents that differ in observation space, availability of short-term memory, and the employed LLM. The validation highlights the difficulty of the benchmark, with the best-performing agents achieving task completion rates below 65\% in the task categories cheapest product search and vague product search.
\end{abstract}

%%
%% The code below is generated by the tool at http://dl.acm.org/ccs.cfm.
%% Please copy and paste the code instead of the example below.
%%
\begin{CCSXML}
<ccs2012>
   <concept>
       <concept_id>10010147.10010257.10010258.10010260</concept_id>
       <concept_desc>Computing methodologies~Intelligent agents</concept_desc>
       <concept_significance>500</concept_significance>
   </concept>
   <concept>
       <concept_id>10002951.10003317.10003359.10003361</concept_id>
       <concept_desc>Information systems~Web searching and information discovery</concept_desc>
       <concept_significance>400</concept_significance>
   </concept>
</ccs2012>
\end{CCSXML}

\ccsdesc[500]{Computing methodologies~Intelligent agents}
\ccsdesc[400]{Information systems~Web searching and information discovery}

%%
%% Keywords. The author(s) should pick words that accurately describe
%% the work being presented. Separate the keywords with commas.
\keywords{Web agents; Product search; E-commerce; Benchmarking}
%% A "teaser" image appears between the author and affiliation
%% information and the body of the document, and typically spans the
%% page.

%%
%% This command processes the author and affiliation and title
%% information and builds the first part of the formatted document.
\maketitle

\section{Introduction}
\label{sec:introduction}

The emergence of large language models (LLMs) and multi-modal agents based on these models has sparked renewed interest in building agents that can browse the World Wide Web and execute complex tasks in this environment~\cite{ning2025survey,ferragLLMReasoningAutonomous2025,Wang_Survey_Agents_2024}. A number of benchmarks have been proposed to evaluate Web agents for online shopping~\cite{yaoWebShop2022,gargREALBenchmarkingAutonomous2025,wangShoppingBenchRealWorldIntentGrounded2025,lyu2025deepshopbenchmark,KimAgenticShopWWW2026,shopperbench2026,AmazonBench2026} as well as on broader ranges of tasks including online shopping~\cite{deng2023mind2web,zhouWebArenaRealisticWeb2023}. These benchmarks either evaluate agents online on the live Web~\cite{deng2023mind2web,lyu2025deepshopbenchmark,touShoppingCompAreLLMs2026,KimAgenticShopWWW2026,AmazonBench2026} or simulate online shops~\cite{yaoWebShop2022,gargREALBenchmarkingAutonomous2025,wangShoppingBenchRealWorldIntentGrounded2025,zhouWebArenaRealisticWeb2023,shopperbench2026}. 
Simulating a web environment allows the exact reproducibility of evaluation results and the comparison of agents using exactly the same environment. The existing benchmarks that simulate online shops~\cite{yaoWebShop2022,gargREALBenchmarkingAutonomous2025,wangShoppingBenchRealWorldIntentGrounded2025,zhouWebArenaRealisticWeb2023,shopperbench2026} all simulate tasks in single shops containing offers from a single source. What is missing are comparison shopping benchmarks in a reproducible environment, which require searching and comparing heterogeneous product descriptions and prices across multiple shops. 

To fill this gap, we introduce WebMall, a multi-shop e-commerce benchmark. WebMall is the first simulated environment benchmark with tasks requiring navigation of multiple web shops to retrieve and aggregate information given various levels of user-query specificity, as well as performing actions such as adding items to carts and finalizing a purchase by checking out. 
Compared to existing simulated environment benchmarks, WebMall shops contain more heterogeneous product descriptions, as they are populated with offers extracted from a wide range of different real-world shops. 
In contrast to existing simulated environment benchmarks, the WebMall task set goes beyond searching for specific products and also includes searches with vague user requirements, price comparisons, and searches for compatible products or cheaper substitute products.
Compared to interacting with a single shop, visiting multiple shops is challenging for Web agents as it results in longer navigation trajectories and requires agents to remember relevant products over longer sequences of interaction steps. 

The WebMall environment comprises four simulated online shops, as well as a set of 91 tasks across 11 task categories that require cross-shop navigation and search. Each task in the benchmark is defined by a natural language instruction and an expected result, e.g., the URLs of the relevant product offers for a given query. The tasks are intentionally designed to require agents to visit multiple shops, search for items, compare offers, rank them by price,  add products to the shopping cart, and proceed to checkout.

In order to validate the level of difficulty of the WebMall benchmark for evaluating the effectiveness and efficiency of web agents for online shopping retrieval tasks, we perform a series of baseline experiments with eight different agent configurations using the Browsergym/AgentLab framework~\cite{chezellesBrowserGymEcosystemWeb2024}. The agents distinguish themselves along three dimensions: (i) observation space (accessibility tree and/or screenshots), (ii) availability of persistent short-term memory, and (iii) internal LLM. 

Our baseline experiments show that the benchmark is challenging for the LLMs GPT-5.4 and Qwen 3.6 Plus. The best configuration achieves completion rates below 65\% in the task categories of cheapest product search and vague product search. We find that using the accessibility tree is most important for successful navigation and achieving high task completion rates, whereas screenshots can be a helpful addition in some situations but cannot replace the structured information found in accessibility trees. We further observe large differences in token consumption, runtime, and API cost between agents using GPT-5.4 and agents using Qwen 3.6 Plus. The absolute token consumption and runtime per task proved to be quite high for both models due to WebMall requiring agents to interact with multiple websites.

The contributions of this paper are:
\begin{enumerate}
    \item \textit{WebMall}, a novel simulated environment benchmark for evaluating web agents, consisting of four e-shops and a challenging comparison shopping task set that covers tasks requiring navigation, interaction, search and reasoning skills. WebMall distinguishes itself from existing simulated environment benchmarks due to (i) tasks requiring navigation of multiple shops, (ii) the heterogeneity of product offers and navigation interfaces across shops, and (iii) its increased task complexity, mirroring comparison shopping customer journeys.
    \item An evaluation of eight baseline agents differing in observation space, the use of short-term memory, and the underlying LLM. We analyze completion rates, precision, recall, F1 scores, token usage, runtime, and cost.
\end{enumerate}

\textbf{Resource Availability:} The WebMall benchmark includes (i) the dockerized multi-shop environment, (ii) the 91-task suite across 11 categories, (iii) the evaluation harness, and (iv) the baseline agents. All artifacts are found on GitHub\footnote{\url{https://github.com/wbsg-uni-mannheim/WebMall}}. After cloning the repository, the two-command setup\footnote{\url{https://github.com/wbsg-uni-mannheim/WebMall/blob/main/docker_all/README.md}} automatically downloads all files, configures the services, and launches the four shops along with their databases and Elasticsearch instances.
\section{The WebMall Environment}
\label{sec:webmall-environment}

 The \textit{WebMall} benchmark consists of four electronics-focused online shops, each hosting a distinct set of product offers. To populate the shops with real-world product offers, we used the WDC Extraction of the October 2024 Common Crawl\footnote{\url{https://webdatacommons.org/structureddata/2024-12/stats/schema_org_subsets.html}}. The extraction files contain product offers from thousands of real-world e-shops that mark up product offers on their websites using the schema.org\footnote{\url{https://schema.org/}} vocabulary. In addition to the four shops, the benchmark environment contains a \textit{solution website} that agents must use to submit their solutions to the tasks described in Section~\ref{sec:taskset} or indicate that they have finished the task if no solution submission is required.

\subparagraph{\textbf{Webmall Shops:}} 

\begin{figure}[h]
  \centering
  \includegraphics[width=\linewidth]{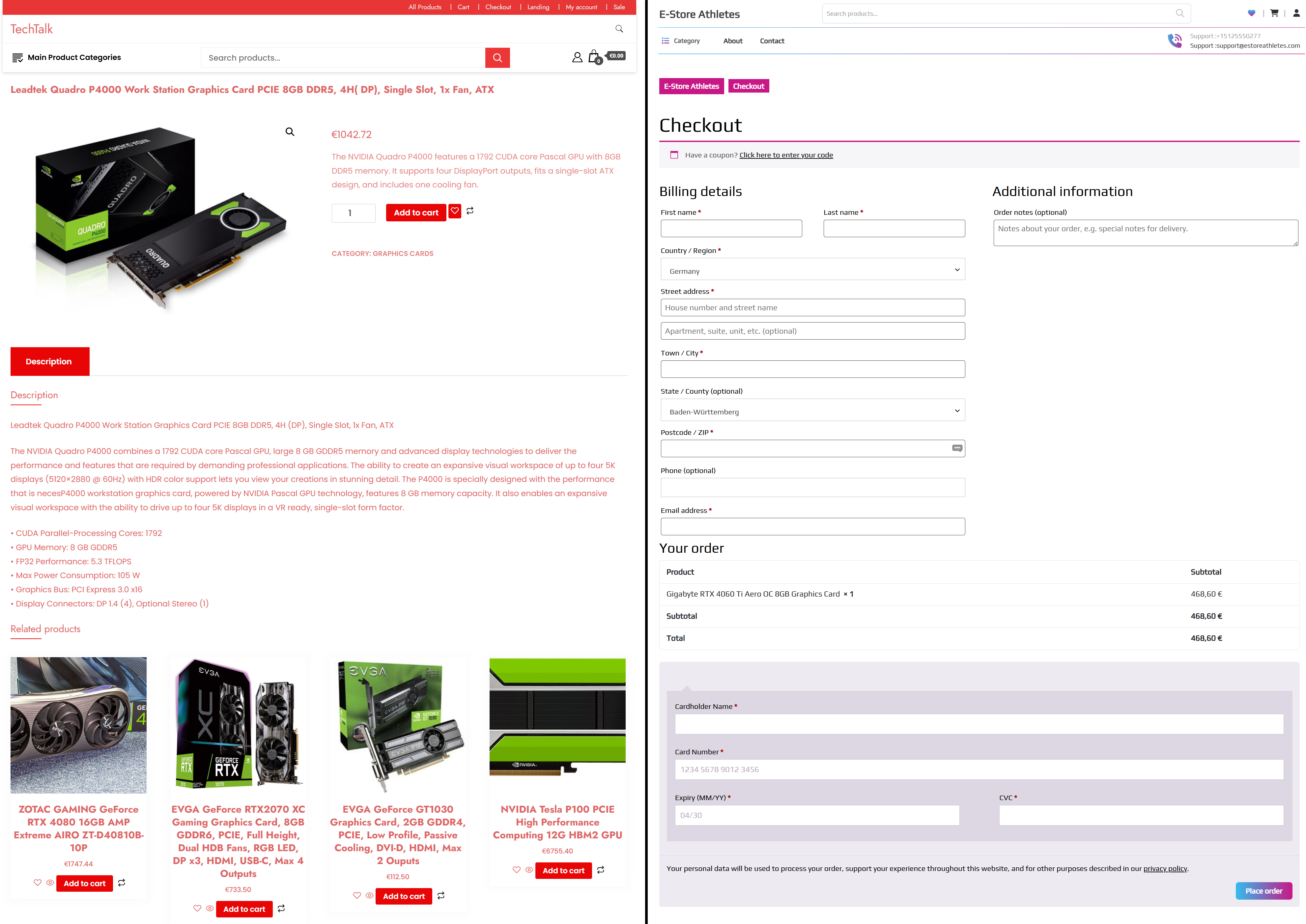}
  \caption{Product detail page (left) and checkout page (right) in two of the WebMall stores.}
  \Description{Left picture shows the detail page of a graphics card with title, description, price, add to cart button and related products. Right picture shows the checkout page with address fields and credit card information fields to be filled.}
  \label{fig:page-example}
\end{figure}

The four shops of WebMall are created using WordPress and are locally hostable with Docker containers. We selected four free-to-use templates from the WooCommerce\footnote{\url{https://woocommerce.com/}} marketplace to implement the shops. All four shops expose heterogeneous interfaces and are visually distinct from one another. Each shop contains a shopping cart, checkout functionality, a search bar, a category drop-down with heterogeneous category trees across the shops, and product detail pages for navigating the shop and finding relevant product offers. Figure~\ref{fig:page-example} shows a product detail and checkout page of two of the four shops.

\begin{table*}[]
\centering
\caption{Product distribution across the four shops.}
\label{tab:product_distribution_across_all_shops}
\resizebox{0.8\textwidth}{!}{%
\begin{tabular}{@{}lcccccccccc@{}}
\toprule
Product Category & \multicolumn{2}{c}{Overall Total} & \multicolumn{2}{c}{Shop 1} & \multicolumn{2}{c}{Shop 2} & \multicolumn{2}{c}{Shop 3} & \multicolumn{2}{c}{Shop 4}\\
 & Offers & \% & Offers & \% & Offers & \% & Offers & \% & Offers & \%\\
\midrule
PC Components & 1,477 & 33.4 & 348 & 30.2 & 369 & 33.7 & 430 & 37.2 & 330 & 32.4\\
PC Peripherals & 1,388 & 31.4 & 432 & 37.5 & 255 & 23.3 & 336 & 29.1 & 365 & 35.8\\
Other Electronics & 1,556 & 35.2 & 370 & 32.3 & 471 & 43.0 & 390 & 33.7 & 325 & 31.9\\
Total & 4,421 & 100.0 & 1,150 & 100.0 & 1,095 & 100.0 & 1,156 & 100.0 & 1,020 & 100.0\\
\bottomrule
\end{tabular}%
}
\end{table*}

\subparagraph{\textbf{Product Offer Collection:}} 

For the selection of product offers to be presented in the four online shops, several constraints must be fulfilled. We filter the set of product offers in the WDC extraction, keeping only those that contain the following schema.org properties: \textit{title}, \textit{description}, \textit{price}, and \textit{priceCurrency}. Afterwards, we deduplicate the filtered product offers by removing exact duplicates based on the combination of all four attributes. WebMall is designed as an English language benchmark. As the extraction files contain product offers in many different languages, we apply the fastText\footnote{\url{https://fasttext.cc/docs/en/language-identification.html}} language classification model to the titles and descriptions of the offers to filter for English offers only. A subset of the remaining product offers has schema.org annotations for globally unique product identifiers, such as \textit{GTIN} or \textit{MPN} numbers. We use these identifiers to group product offers referring to the same real-world product into clusters. These clusters facilitate the selection of product offers for the same product when later distributing the offers across the four online shops and creating the shopping tasks. 

 After the filtering and clustering step, we manually select product offers for electronics products  while creating the tasks described in Section~\ref{sec:taskset} and distribute them across the shops accordingly. To mimic a realistic comparison shopping scenario, we partition the products across the four shops so that each shop contains a mix of \textit{PC components}, \textit{PC peripherals}, and \textit{other electronics}. The \textit{other electronics} category contains devices such as digital cameras, smartphones, and smartwatches, along with related accessories. Table~\ref{tab:product_distribution_across_all_shops} shows the resulting distribution of the collected 4,421 product offers across the shops by category. The category trees in each shop are different, simulating the heterogeneity found in real e-shops, and are manually created by the authors.

Across all four WebMall shops, the product offers feature heterogeneous titles and descriptions. Titles range from 6 to 264 characters, with a median length of 69 and an average of 76.4. The middle 50\% of titles fall between 45 and 100 characters, with 90\% being shorter than 135 characters. Descriptions are substantially longer, spanning 15 to over 14,000 characters, with a median of 573 and an average of about 1,059. The middle 50\% of descriptions range from 339 to 1,330 characters, and 90\% are shorter than 2,542 characters.

\section{The WebMall Task Set}
\label{sec:taskset}

The WebMall tasks require agents to search for products across multiple shops.
Prior work in simulated shopping environments~\cite{yaoWebShop2022,zhouWebArenaRealisticWeb2023,gargREALBenchmarkingAutonomous2025,wangShoppingBenchRealWorldIntentGrounded2025} largely focuses on interactions within a single store. In contrast, real-world online shopping often involves product search across multiple shops exhibiting heterogeneous product descriptions, category structures, and pricing. The searches include both well-specified and vague queries.

Each WebMall task is expressed as a natural-language query. The retrieval units are product offer pages, distributed across four shops. Relevance is provided as a set of solution product offer URLs, enabling evaluation of both set retrieval (high recall of all relevant offers) and specific selection/ranking (choosing the cheapest relevant offer under constraints). The setting of WebMall offers interactive search, where retrieval decisions are combined with actions such as navigation, adding product offers to the shopping cart and form-filling checkout steps.

To cover a broad range of e-commerce retrieval problems, WebMall includes tasks spanning exact product search, attribute- and constraint-based retrieval, search under vague requirements, price-aware ranking, substitute and compatibility retrieval, and end-to-end workflows. The heterogeneity across shop environments makes the benchmark suitable for assessing robustness of retrieval performance and decision-making capabilities of web agents.

\begin{table*}[]
\centering
\caption{Overview of the WebMall task set.}
\label{tab:task_categories_overview}
\resizebox{\textwidth}{!}{%
\begin{tabular}{@{}llcl@{}}
\toprule
Task Group & Task Category & Count & Example Tasks\\
\midrule
\multirow{2}{*}{Specific Product Search} & Find Specific Product & 12 & Find all offers for the AMD Ryzen 9 5900X.\\
 & Products Fulfilling Specific Req. & 11 & Find all offers for orange straps that fit with the Apple Watch Series 6.\\
\midrule
\multirow{3}{*}{Vague Product Search} & Products Satisfying Vague Req. & 8 & Find all offers for compact keyboards that are best suited for working with a laptop remotely.\\
 & Find Substitutes & 6 & Find the cheapest alternative for this item: \{PRODUCT\_URL\}.\\
 & Find Compatible Products & 5 & Find all offers for compatible CPUs for this motherboard: \{PRODUCT\_URL\}.\\
\midrule
\multirow{3}{*}{Cheapest Product Search} & Find Cheapest Offer & 10 & Find the cheapest offer for the Samsung Galaxy S24 Plus.\\
 & Cheapest Offer Specific Req. & 10 & Find the cheapest offer for a new Xbox gaming console with at least 512gb disk space in white.\\
 & Cheapest Offer Vague Req. & 6 & Find the cheapest offers for each model of mid-tier nVidia gaming GPUs in the 4000 series.\\
\midrule
\multirow{2}{*}{Action \& Transaction} & Add to Cart & 7 & Find all offers for the Asus DUAL RTX4070 SUPER OC White and add them to the shopping cart.\\
 & Checkout & 8 & Add the product on page \{PRODUCT\_URL\} to the shopping cart and complete the checkout process.\\
\midrule
End to End & End To End & 8 & Find the cheapest offer for the Asrock B550 PHANTOM GAMING 4 and purchase it.\\
\bottomrule
\end{tabular}%
}
\end{table*}

The WebMall task set contains 91 tasks across 11 task categories, grouped into five task groups. Tasks consist of a natural-language instruction and a set of solution URLs. Table~\ref{tab:task_categories_overview} summarizes the categories and provides an example for each. All tasks were created and cross-checked by domain experts. Tasks and solutions were tested and refined over multiple baseline agent runs and human trials before finalizing the task set.

\textbf{Specific Product Search} tasks correspond to exact product queries and attribute-constrained retrieval, where users precisely specify the target product. This includes \textit{Find Specific Product} tasks (retrieve all offers for a named product across all shops) and \textit{Products Fulfilling Specific Requirements} tasks (retrieve all offers satisfying explicit attribute constraints such as display size).

\textbf{Vague Product Search} tasks capture retrieval under vague user specifications and present query interpretation challenges. \textit{Products Satisfying Vague Requirements} require mapping vague natural-language preferences to product attributes and retrieving the relevant set of product offers. \textit{Find Substitutes} tasks target alternative retrieval under constraints (e.g., cheaper but functionally similar), and \textit{Find Compatible Products} tasks target complementary retrieval where relevance depends on explicit compatibility constraints (e.g., socket/chipset/form factor).

\textbf{Cheapest Product Search} tasks evaluate price-based ranking and selection after relevance and constraints are satisfied. They include \textit{Find Cheapest Offer} tasks (minimal price among relevant offers for a named product across shops), \textit{Cheapest Offer with Specific Requirements} tasks (constraint-based retrieval followed by ranking and selection), and \textit{Cheapest Offer with Vague Requirements} tasks (interpreting vague user specifications, retrieval, ranking, and selection).

\textbf{Action \& Transaction} tasks include \textit{Add To Cart} (adding offers for specific products to the cart) and \textit{Checkout} tasks (adding an offer to the cart and proceeding through checkout, including providing shipping and billing details via HTML forms). 

\textbf{End-to-End} tasks combine price-aware retrieval and selection with cart and checkout tasks into a single interactive workflow, representing full-session shopping behavior from search to purchase completion.
\section{Experimental Evaluation}
\label{sec:experiments}

To validate the usefulness of the WebMall benchmark for evaluating the effectiveness and efficiency of web agents, we experiment with different agent design choices on \textit{WebMall} using the Browsergym and AgentLab frameworks~\cite{chezellesBrowserGymEcosystemWeb2024}. 
The Browsergym framework offers a set of common tools for agents, such as web browsing capabilities using the Python playwright library, experimental framing, and result/trace tracking for agents that work with any API-based hosted LLM. This functionality allows users to easily run and compare agents on various benchmarks. The AgentLab library that integrates with Browsergym can be used to configure and run more sophisticated agents by providing API-based LLMs with a set of capabilities, such as using accessibility trees, screenshots, short-term memory, and custom instruction prompts. 

We experiment on WebMall using Browsergym/AgentLab with eight baseline agent configurations that vary along three axes:

\begin{table*}[]
\centering
\caption{Completion rates and F1 scores aggregated by task group. Best results are shown in bold, second best underlined.}
\label{tab:completion_rates_and_f1_score_by_aggregate_task_groups}
\resizebox{\textwidth}{!}{%
\begin{tabular}{@{}llcccccccccccccccc@{}}
\toprule
\multirow{2}{*}{Model} & \multirow{2}{*}{Task Group} & \multicolumn{4}{c}{AX-Tree} & \multicolumn{4}{c}{AX-Tree + Memory} & \multicolumn{4}{c}{AX-Tree + Vision} & \multicolumn{4}{c}{Vision}\\
 &  & CR (\%) & P (\%) & R (\%) & F1 (\%) & CR (\%) & P (\%) & R (\%) & F1 (\%) & CR (\%) & P (\%) & R (\%) & F1 (\%) & CR (\%) & P (\%) & R (\%) & F1 (\%)\\
\midrule
\multirow{5}{*}{GPT-5.4} & Specific Product Search & \textbf{69.32} & \textbf{90.38} & \textbf{79.87} & \textbf{84.78} & 65.15 & 82.58 & 71.50 & 76.60 & 52.27 & \underline{84.66} & 67.87 & 75.34 & 64.39 & 77.27 & 68.47 & 72.61\\
 & Cheapest Product Search & 54.44 & \underline{68.89} & \underline{62.59} & \underline{65.44} & \textbf{63.33} & \textbf{70.28} & \textbf{66.67} & \textbf{68.32} & 43.33 & 54.17 & 54.72 & 54.44 & 36.67 & 36.67 & 36.67 & 36.67\\
 & Vague Product Search & \textbf{64.44} & \textbf{87.57} & \textbf{82.92} & \textbf{85.13} & \underline{54.72} & 83.33 & 72.50 & 77.38 & 50.56 & \underline{84.05} & \underline{76.23} & \underline{79.82} & 35.83 & 57.30 & 53.75 & 55.44\\
 & Action \& Transaction & \textbf{100.00} & \underline{100.00} & \textbf{100.00} & \textbf{100.00} & \underline{100.00} & 100.00 & \underline{100.00} & \underline{100.00} & 100.00 & 100.00 & 100.00 & 100.00 & 79.46 & 93.75 & 86.61 & 89.90\\
 & End To End & 62.50 & 75.00 & 68.75 & 71.74 & 75.00 & \textbf{87.50} & \textbf{81.25} & \textbf{84.26} & 75.00 & \underline{87.50} & \underline{81.25} & \underline{84.26} & 25.00 & 37.50 & 31.25 & 34.09\\
\midrule
\multirow{5}{*}{Qwen 3.6 Plus} & Specific Product Search & 51.52 & 80.43 & 68.58 & 74.03 & \underline{68.94} & 81.82 & \underline{74.23} & \underline{77.84} & 39.39 & 73.11 & 55.51 & 62.80 & 0.00 & 8.33 & 4.86 & 6.14\\
 & Cheapest Product Search & \underline{54.44} & 65.56 & 59.07 & 61.76 & 54.44 & 63.57 & 61.30 & 62.37 & 16.67 & 16.67 & 16.67 & 16.67 & 3.33 & 5.00 & 6.67 & 5.56\\
 & Vague Product Search & 38.33 & 56.81 & 55.14 & 55.60 & 43.89 & 59.37 & 59.38 & 59.33 & 28.61 & 35.56 & 32.78 & 33.97 & 0.00 & 0.00 & 0.00 & 0.00\\
 & Action \& Transaction & 78.57 & 92.86 & 85.71 & 88.96 & 85.71 & \textbf{100.00} & 92.86 & 96.15 & 66.96 & 81.25 & 74.11 & 77.40 & 0.00 & 28.57 & 14.29 & 19.05\\
 & End To End & \textbf{75.00} & 75.00 & 75.00 & 75.00 & \underline{75.00} & 75.00 & 75.00 & 75.00 & 12.50 & 12.50 & 12.50 & 12.50 & 0.00 & 0.00 & 0.00 & 0.00\\
\bottomrule
\end{tabular}%
}
\end{table*}
\begin{table*}[]
\centering
\caption{Steps, token usage, runtime, and cost across all tasks. Best results are shown in bold, second best underlined.}
\label{tab:token_usage_cost_and_runtime_across_all_tasks}
\resizebox{0.8\textwidth}{!}{%
\begin{tabular}{@{}llccccc@{}}
\toprule
Model & Observation Space & Avg. Steps & Avg. Input Tokens & Avg. Output Tokens & Avg. Runtime & Avg. Cost\\
\midrule
\multirow{4}{*}{GPT-5.4} & AX-Tree & 25.11 & \underline{148,250} & \underline{2,178} & \textbf{128s} & \$0.40\\
 & AX-Tree + Memory & \textbf{23.26} & 187,156 & 6,362 & 165s & \$0.56\\
 & AX-Tree + Vision & \underline{23.36} & 162,041 & \textbf{2,165} & \underline{141s} & \$0.44\\
 & Vision & 31.35 & \textbf{119,866} & 2,862 & 166s & \$0.34\\
\midrule
\multirow{4}{*}{Qwen 3.6 Plus} & AX-Tree & 26.97 & 242,900 & 5,965 & 234s & \$\underline{0.09}\\
 & AX-Tree + Memory & 27.12 & 319,728 & 10,625 & 321s & \$0.12\\
 & AX-Tree + Vision & 29.78 & 258,221 & 4,753 & 241s & \$0.09\\
 & Vision & 49.17 & 192,212 & 5,958 & 381s & \textbf{\$0.07}\\
\bottomrule
\end{tabular}%
}
\end{table*}

\begin{itemize}
    \item \textbf{Observation space:} Agents may perceive the current webpage through an HTML accessibility tree (denoted \textit{AX-Tree}), a screenshot of the viewport of the webpage (\textit{Vision}), or a combination of both (\textit{AX-Tree+Vision}). The accessibility tree provides structural information, such as input fields and their labels, while the screenshot can capture more visual cues, such as product images and the visual layout of the current page.
    \item \textbf{Memory:} With \textit{memory}, agents can maintain a persistent memory scratchpad across steps that allows them to store and filter discovered information, such as found product offers and their URLs. In \textit{no-memory} configurations, agents rely solely on an action history and their thoughts at each step.
    \item \textbf{Large language model:} We evaluate \textit{GPT-5.4} (gpt-5.4-2026-03-05) via the OpenAI API and \textit{Qwen 3.6 Plus} via the OpenRouter API as AgentLab backends, contrasting a state-of-the-art proprietary model with an open-weight alternative. Additional results for the older models GPT4.1 and Claude Sonnet 4 are available on the project GitHub repository.

\end{itemize}

For each task, we allow the agent up to 50 steps to complete it. These steps are a sequence of actions such as \textit{go to page}, \textit{click}, \textit{fill text} and \textit{scroll}. This action set is defined by AgentLab and passed to the agent at each step. An example of a full agent trace and the history passed to the agent at the final step of an example task can be found on GitHub\footnote{\url{https://github.com/wbsg-uni-mannheim/WebMall/blob/main/examples/task_message.txt}}.

\subsection{Effectiveness Analysis}
\label{subsec:effectiveness-analysis}

We measure agent effectiveness using four task-level metrics: task completion rate (CR), precision (P), recall (R), and F1.
For each task~$i$, let $G_i$ be the gold answer set defined by the benchmark and $A_i$ the answer set produced by the agent.
For product-search tasks, $G_i$ and $A_i$ contain product URLs. 
For add-to-cart and checkout tasks, $G_i$ is the set of offer URLs that are required to be in the cart, respectively to have been checked out.

The task completion rate is a strict exact-match indicator,
so a task is counted as complete only when the agent returns every relevant item and no additional irrelevant items.
To capture partial completion, we compute per-task precision and recall on the answer sets and aggregate by macro-averaging over the tasks in each task category.

Table~\ref{tab:completion_rates_and_f1_score_by_aggregate_task_groups} shows the results of the WebMall validation experiments. The results are averaged per task group, due to space constraints.
More detailed task category-level results are found on the project website\footnote{\url{https://wbsg-uni-mannheim.github.io/WebMall/}}. 

\subparagraph{\textbf{Specific Product Search:}} The two backend models diverge on this task group: GPT-5.4 reaches its best result with the accessibility tree alone (69\% CR and 85\% F1), while adding memory or screenshots harms the performance. Qwen 3.6 Plus, in contrast, benefits from memory, climbing from 52\% to 69\% CR and from 74\% to 78\% F1 when the AX-Tree is paired with the short-term memory scratchpad. The traces show that GPT-5.4 converges on the correct product detail page faster, while Qwen visits more candidate pages before arriving at the right one (see average step counts in Table~\ref{tab:token_usage_cost_and_runtime_across_all_tasks}). The Qwen model profits from explicit memory because it helps keep the model focused across these longer trajectories.

\subparagraph{\textbf{Cheapest Product Search:}} For these tasks, both models benefit from adding memory in addition to using the AX-Tree. For GPT-5.4, the AX-Tree+Memory configuration attains the highest aggregate scores in this task group (63\% CR, 68\% F1), whereas the Qwen 3.6 Plus model achieves best results (54\% CR and 62\% F1) also for AX-Tree+Memory agents. The best CR and F1 values are lower than for \textit{Specific Product Search} for both backend models. This suggests that price comparison adds a non-trivial challenge to agents on top of product search.

\subparagraph{\textbf{Vague Product Search:}} For this task group, the effect of memory is model-dependent: GPT 5.4 achieves its best results (64\% CR and 85\% F1) only using the accessibility tree, while Qwen performs best (44\% CR and 59\% F1) using AX-Tree+Memory. Adding screenshots on top of the accessibility tree does not help either model, as performance is lower than their AX-only results. GPT-5.4 achieves better scores overall in this task group, indicating stronger reasoning abilities regarding vague requirements of a user compared to Qwen 3.6 Plus.

\subparagraph{\textbf{Action \& Transaction:}} GPT-5.4 reliably completes all tasks across every AX-based observation space, with all three AX-Tree configurations reaching a completion rate of 100\%. Qwen 3.6 Plus achieves its best completion rate of 85\% with AX-Tree+Memory. Vision-only agents perform worse than AX-based agents in this group due to the missing structural grounding, and for Qwen 3.6 Plus they generally fail to solve tasks independent of task group (see discussion below).

\subparagraph{\textbf{End To End:}} GPT-5.4 reaches top performance on full shopping workflows with specific user queries with both AX-Tree+Memory and AX-Tree+Vision, achieving a completion rate of 75\% and an F1 of 84\%. The remaining end-to-end task failures for GPT-5.4 are the same for both configurations, indicating that neither has an advantage over the other for a specific task from this group. Qwen 3.6 Plus matches the 75\% completion rate with the AX-Tree-only and AX-Tree+Memory configurations, but at a lower F1 of 75\%, which means that GPT-5.4 performs better on the remaining unfulfilled tasks, but neither model manages to fully complete them.

Across all task groups, agents using Vision in addition to the AX-Tree can improve results on \textit{End To End} tasks for GPT-5.4, but do not yield consistent gains and do in most cases degrade performance, such as for \textit{Cheapest Product Search} and \textit{Vague Product Search}. Agents relying exclusively on screenshots are substantially weaker than AX-based agents and, in the case of Qwen 3.6 Plus, fail entirely on every task group. Trace-level inspection shows that Qwen 3.6 Plus Vision agents fail by repeatedly emitting the same click action against an invisible or stale element until either the AgentLab harness stops the execution due to repeated failed attempts or the step budget is exhausted, indicating that the model cannot maintain a stable mapping from the rendered screenshot to the navigation element bid index across page re-renders. This demonstrates that visual inputs without structural guidance from the AX-Tree are insufficient for task completion when the underlying model is not strong enough to recover from grounding mismatches.

\begin{figure*}[h]
  \centering
  \includegraphics[width=0.8\textwidth]{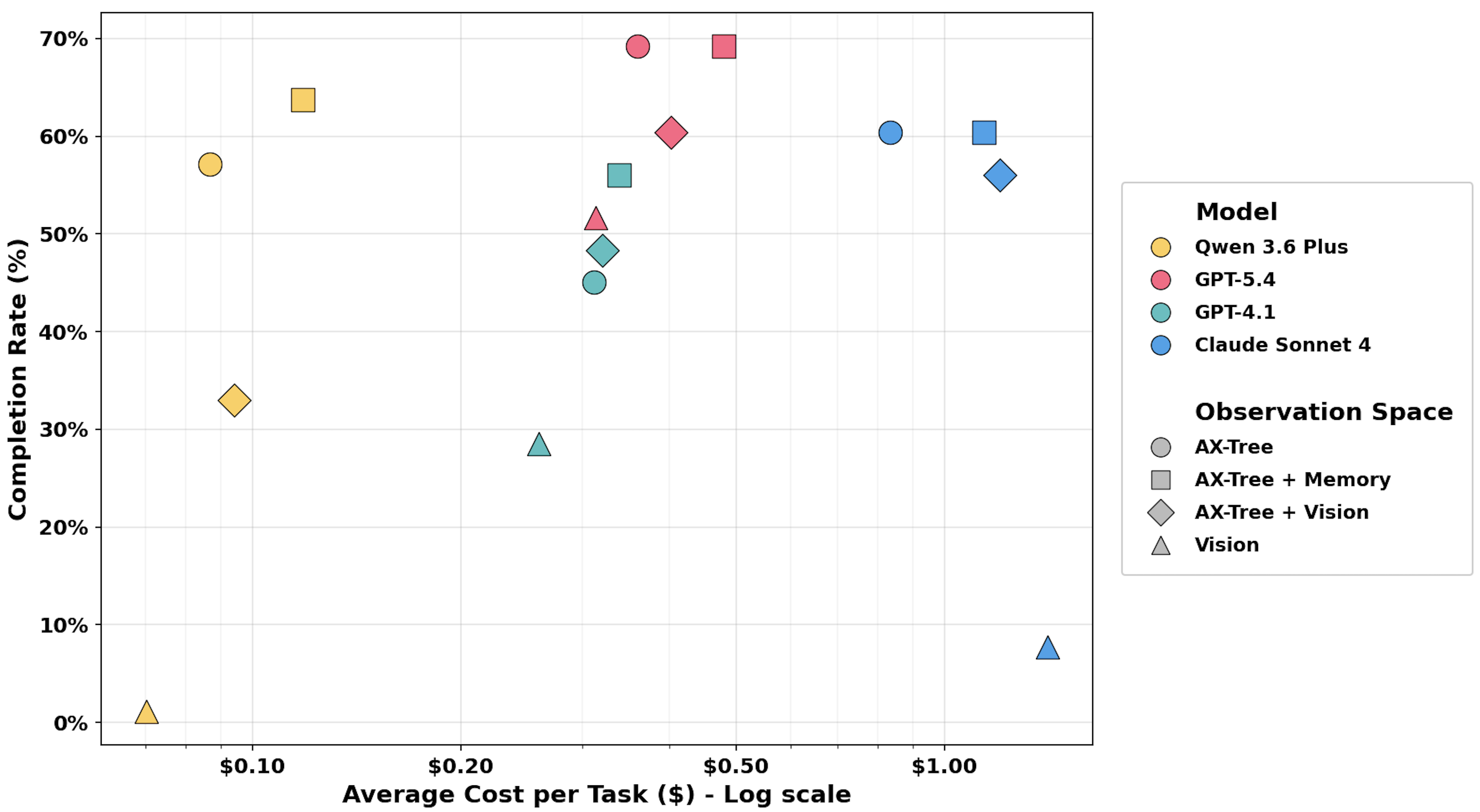}
  \caption{Cost versus task completion rate across all tasks.}
  \Description{}
  \label{fig:costvsperformance}
\end{figure*}

\subsection{Efficiency Analysis}
\label{subsec:efficiency-analysis}
Efficiency is a key concern for real-world agent deployment. Table~\ref{tab:token_usage_cost_and_runtime_across_all_tasks} shows an overview of average token usage, API cost, and runtime for each agent configuration across all tasks. 

\subparagraph{\textbf{Token Usage:}} Agents based on Qwen 3.6 Plus consume substantially more tokens per task than GPT-5.4 configurations, with average input token counts between 192,000 and 320,000 compared to about 120,000-187,000 for GPT-5.4. This difference is due to the tokenizer of the Qwen 3.6 Plus model requiring more tokens for encoding an AX-Tree compared to GPT-5.4. This is likely an artifact of the vocabulary generation process of both tokenizers. Additionally, the increased step count for the Qwen 3.6 Plus models further increase the average token counts as more steps result in a longer action history. Adding screenshots tends to increase token usage for Qwen due to an increase in average steps for solving a task, whereas GPT-5.4 Vision-only agents use the fewest input tokens among the GPT-5.4 variants. The latter is due to the encoded screenshots being 3-5x smaller with regards to the required tokens compared to the average AX-Tree in WebMall. 
For both models, screenshot-only agents require the highest average number of steps (31.35 for GPT-5.4 Vision and 49.17 for Qwen Vision) mainly due to getting lost during navigation. Memory-based configurations slightly reduce the average number of steps for GPT-5.4 (from 25.11 to 23.26), while leaving Qwen essentially unchanged, but in either case this is not enough to offset the additional memory context with an overall increased total input and output tokens compared to using only the AX-Tree.

\subparagraph{\textbf{Runtime:}} Across all tasks, GPT-5.4 configurations complete a task in around 2.1-2.8 minutes on average, whereas Qwen 3.6 Plus agents work between 3.9 and 6.4 minutes per task, with the longest runtimes observed for screenshot-only configurations. This difference mirrors the higher token usage and step counts of Qwen models and screenshot-heavy setups. Together with the completion rates and F1 scores reported in Table~\ref{tab:completion_rates_and_f1_score_by_aggregate_task_groups}, this suggests that GPT-5.4 is the more efficient choice when runtime is the bottleneck.

\subparagraph{\textbf{API Usage Fees:}} The cost per task are determined by token usage and model pricing. During our experiments, GPT-5.4 had a cost of \$2.50 per 1M input tokens and \$15.00 per 1M output tokens. Qwen 3.6 Plus had a cost of \$0.325 per 1M input tokens and \$1.95 per 1M output tokens. Across all tasks, Qwen 3.6 Plus configurations are by far the most cost-effective, with average costs between \$0.07 and \$0.12 per task, whereas GPT-5.4 agents range from \$0.34 (Vision) up to \$0.56 (AX-Tree+Memory). 
Overall, there is a clear trade-off between performance and efficiency: more sophisticated and modality-rich agent configurations, particularly those based on GPT-5.4 with screenshots and memory, tend to achieve higher success rates in challenging task categories, but at a substantial increase in token usage, runtime, and cost relative to the open-weight Qwen 3.6 Plus alternative. Figure~\ref{fig:costvsperformance} shows a comparison of the average cost per task and the completion rate for all models, including the older GPT-4.1 and Claude Sonnet 4. The full results per task group for the latter two models can be found on the WebMall webpage. The figure shows that the completion rate has increased by more than 10\% across all observation spaces with GPT-5.4 compared to GPT-4.1, although the cost also increased. Compared to the expensive Claude Sonnet 4 model, the newer Qwen 3.6 Plus is a significantly cheaper model achieving similar completion rates.

\section{Related Work}
\label{sec:relatedwork}

In the following, we compare the WebMall benchmark with other benchmarks for evaluating web agents.
An early benchmark in the area of online shopping is WebShop~\cite{yaoWebShop2022}, which simulates a single e-shop populated with over one million product offers from Amazon. Mind2Web~\cite{deng2023mind2web} uses snapshots of real websites to simulate a Web environment and covers various tasks, including online shopping. Each Mind2Web shopping task is limited to a single website, meaning that the agents do not need to perform cross-shop comparisons and that the resulting trajectories are shorter than the trajectories of the WebMall agents. WebArena~\cite{zhouWebArenaRealisticWeb2023} simulates multiple websites spanning domains such as e-commerce, social media, and productivity. However, its shopping tasks are limited to a single e-shop and focus on the administration of the shop and the generation of statistics about sales. The REAL benchmark~\cite{gargREALBenchmarkingAutonomous2025} spans various types of tasks, including shopping tasks in a single-shop environment, such as product search, managing a shopping cart, and completing a checkout process. ShoppingBench~\cite{wangShoppingBenchRealWorldIntentGrounded2025} also simulates a single shop and requires agents to perform task such as searching for products, using vouchers, and sticking to a given budget.

Benchmarks that evaluate agents online on the live Web include Online-Mind2Web~\cite{xueIllusionProgressAssessing2025}, an extension of the original Mind2Web benchmark to the live Web, BrowseComp~\cite{weiBrowseCompSimpleChallenging2025}, which consists of purposefully difficult tasks requiring reasoning and web search, DeepShop~\cite{lyu2025deepshopbenchmark}, which features complex product search queries, and ShoppingComp~\cite{touShoppingCompAreLLMs2026}, which also features complex product search queries including safety-critical search constraints. Amazon-Bench~\cite{AmazonBench2026} covers several e-commerce related tasks and puts an additional focus on agent safety. ShopperBench~\cite{shopperbench2026} and AgenticShop~\cite{KimAgenticShopWWW2026} assess the ability of web agents to perform persona-guided shopping.

WebMall distinguishes itself from existing simulated environment benchmarks in the domain of online shopping by being the first benchmark to introduce comparison shopping tasks that require navigation and search of multiple web shops to collect and aggregate information given various levels of user-query specificity. Furthermore, the product offers in WebMall shops are more heterogeneous, as they originate from hundreds of distinct real-world online shops. The tasks in WebMall require longer interaction trajectories than those in WebShop~\cite{yaoWebShop2022} and Mind2Web~\cite{deng2023mind2web}. In contrast to existing simulated environment benchmarks, the WebMall task set goes beyond searching for specific products and also includes searches with vague user requirements, price comparisons, and searches for compatible products or cheaper substitute products. In contrast to live Web benchmarks like Online-Mind2Web~\cite{xueIllusionProgressAssessing2025}, DeepShop~\cite{lyu2025deepshopbenchmark},  ShoppingComp~\cite{touShoppingCompAreLLMs2026}, and AgenticShop~\cite{KimAgenticShopWWW2026}, WebMall confines agents to four specific shops, ensuring that evaluation results are reproducible and enabling the direct comparison of agents within an identical environment.

Further benchmarks for LLM-based agents include AgentBench~\cite{liuAgentBenchEvaluatingLLMs2023}, which extends beyond the Web to databases and operating systems, VisualWebArena~\cite{kohVisualWebArenaEvaluatingMultimodal2024}, and WebChoreArena~\cite{miyaiWebChoreArenaEvaluatingWeb2025}, which focus on visually grounded and memory-intensive tasks, respectively.

Our paper~\cite{SteinerRAGvsMCPvsHTMLWWW2026} builds on WebMall and uses the benchmark to compare the effectiveness and efficiency of different agent interfaces to the Web, including HTML browsing, RAG search, MCP interaction, and NLWeb querying. 

\section{Conclusion}
\label{sec:conclusion}

We presented \textit{WebMall}, the first benchmark for evaluating web agents on e-commerce comparison shopping tasks in a simulated environment consisting of multiple e-shops. The WebMall task set features challenging tasks such as price comparisons and searches for substitute or compatible products. 

Our experiments validated the difficulty of the benchmark, with the best-performing agents achieving task completion rates below 65\% in the task categories of cheapest product search and vague product search. The experiments further showed that structural grounding via the accessibility tree is crucial for successful task completion. While the experiments showed large differences in token consumption, runtime, and API cost between agents using GPT-5.4 and agents using Qwen 3.6 Plus, the absolute token consumption and runtime per task proved to be quite high for both models. This is due to WebMall requiring agents to interact with multiple websites, which leads to long trajectories.

These results highlight the need for agents to be equipped with better product comparison capabilities and for more efficient agent architectures that reduce token consumption without compromising task completion. To that end, we invite the community to use WebMall for research into web agents that can effectively and efficiently perform in multi-shop e-commerce scenarios involving long-running tasks.

%%
%% The next two lines define the bibliography style to be used, and
%% the bibliography file.
\bibliographystyle{ACM-Reference-Format}
\balance
\bibliography{main}

\end{document}